# Automatic liver segmentation method in CT images

Oussema Zayane[1], Besma Jouini[1], Mohamed Ali Mahjoub[2]

[1]University of sousse – Tunisia- oussemazayan@gmail.com

[2]University of Monastir – Tunisia – medalimahjoub@gmail.com

**Abstract** — *The aim of this work is to develop a method for automatic segmentation of the liver based on a priori knowledge of the image, such as location and shape of the liver.*

**Key Words :** Automatic liver segmentation

## 1. INTRODUCTION

Today, the medical image segmentation attracts more and more attention and interest. It provides, automatically, to delimit the internal structures of the patient, these structures can be anatomical (organs) but also pathological (lesions). Automatic segmentation of lesions in a large image database has an obvious interest: it allows the radiologist to assist in diagnosis, by detecting possibly forgotten lesions, and also to accelerate the process of analysis. Therefore the automatic segmentation of the liver plays an important role in the study of liver function and can assist in diagnosis of liver diseases such as steatosis, fibrosis, etc....

In addition, the liver is one of the most important organs of the human body. When it is affected by a tumoral pathology, it is possible to operate it by cutting the sick part. But this cut has to respect rules of volumetrics and very specific vascularization. The medical imaging is then used to detect and visualize the internal structures. These structures do not appear in a single image, but requires several acquisitions which will therefore be compared. The tumoral or hepatic volumetrics when to her is possible only after a stage of segmentation of these images. The objective of the current research works is to achieve the segmentation automatically, to obtain the volume of the liver and its internal structures, in particular the tumors.

The aim of this work is to study the main difficulties in automatic segmentation of the liver, and propose a method to solve them. We will apply this method to CT images: the general principles, however, remain valid for other acquisition techniques (magnetic resonance imaging or MRI, in particular).

The development of the method of automatic segmentation of the liver is based on a priori knowledge of the image, such as location and shape of the liver. Indeed, anatomical point, the liver is located in the upper right quadrant of the abdomen, the right of the stomach and above the intestines. It extends from the abdomen to the thorax. From the perspective of the observer locates the liver is on the left side of the CT image. The liver is the largest organ in the abdomen and the entire human body, we will point out later that this characteristic is very important for the GCC algorithm.

### 1.1 Images to be treated

In the schedules and medical images there are several views and sections guide the anatomical reference system (RAS): it is a terminology based on a set of planes and axes, it is used to indicate the orientation of cuts or views used from Poirier's position, that is when the patient is standing in front of the observer. This is an explanatory diagram:

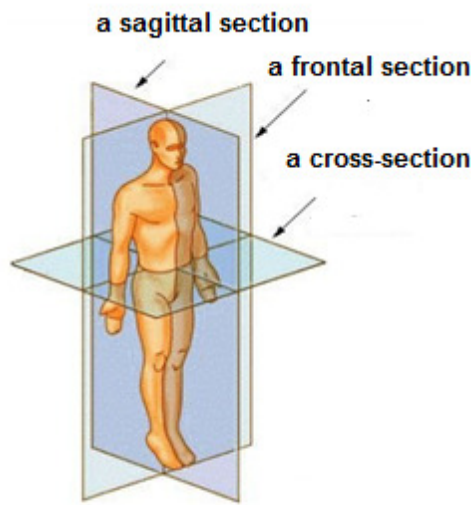

Figure 1: reference position or Troisier position [6]

We are interested in Our project to study the axial cups (cuttings) of the abdomen.

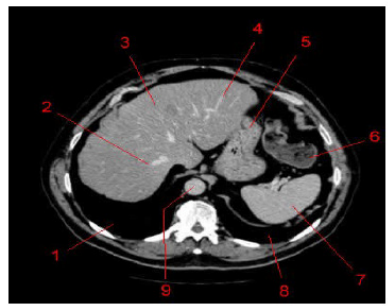

Figure 2: Axial section of the abdomen

1, Right lung. 2, right hepatic vein. 3, Liver. 4, left hepatic vein. 5, Stomach. 6, splenic flexure. 7, Rate. 8, the left lung. 9, Aorta.

## 2. PREVIOUS WORKS

**2.1. The method Max-Flow/Min-Cut [1]**
This is a semi-automatic segmentation of the liver based on graph theory and more specifically on the "Graph Cuts." In this case the problem of segmentation is considered the separation of an image (or volume) into two classes "object" and "bottom". This method is a tool for semi-automatic segmentation that combines interactively some of the voxels of the volume to one of these two classes. This first association serves then as base of training for the final segmentation of the volume.
Like many problems that can be expressed as an optimization of energy, this approach implements an energy minimization method based on partitioning a graph into two subgraphs by cutting minimum capacity.

**2.2. Segmentation based on the deformable models (Snakes) [2]**

This approach of segmentation is based on the deformable models allowing to detect automatically the outlines of the liver in MRI. The methods of segmentation by deformable model (active contours) commonly used in the literature give good results but require manual initialization close to the real contour. In this method the initialization is done in an automatic way.
Automatic initialization is done by studying the histogram and the low processing level salaries. The histogram of the image is modeled by a sum of Gaussian and a Rayleigh curve. The Rayleigh curve represents the noisy background of the image. The Gaussian defines the different classes of grayscale of the image. Various filtering and thresholding are applied successively to obtain a contour similar to that of the liver.
This contour is the deformable model initialization which is then attracted to the boundaries of the form to detect (liver) according to different internal and external forces which force of gradient vector flow (GVF) and a power ball.

**2.3. Histogram Threshold Tail (HTT) Approach [4]**
This method is based on the histogram and thresholding algorithm using HTT, and the estimated MAP (Maximum A Posterior Probability). First we realize the initial detection of the region of interest ROI, then the algorithm will HTT used to remove neighboring abdominal organs. At first, a preprocessing is performed: the ROI is extracted using a multimodal thresholding and binary morphological filter to improve the quality of the segmentation of the liver.

## 3. PRESENTATION OF THE PROPOSED METHOD

As mentioned above, and for the purpose of identification of the liver in CT images, the work involved in this final project study is to develop an automatic segmentation method based on a priori knowledge of the image, as the location and shape of the liver:
The liver is located in the upper right quadrant of the abdomen, right side of the stomach and above the intestines. It extends from the abdomen to the thorax. From the perspective of the observer the liver is located on the left side of the CT image.
The liver is the largest organ in the abdomen and the entire human body, we will point out later that this characteristic is very important for the GCC algorithm. This segmentation will have problems because of the similarity between neighboring pixels of the abdominal organs such as spleen, pancreas, and kidneys. In addition, these problems related to the intensity there are other problems like the choice of operators problem: It is necessary to go through a processing line to correct defects associated with acquiring and improving the quality of the image before performing intelligent analysis of image for recognition of different forms. This processing line will be composed of a number of image processing operators. The choice of these operators and their sequence must be done carefully. In fact, the problem will arise is to use an optimal processing chain, with the least possible redundant operators. This task is far from being easy.

**3.1 The method**
The principle of the proposed method is illustrated in Figure 4. Indeed, the threshold used to separate the pixels of the liver from the rest of the image. In our case, and according to some histograms of the images abdominal CT scan cuts, we see that the histogram of the liver has a shape like that in figure 3. The two lines s1 and s2 are chosen from tests made on a set of CT images X. From these two found values, the thresholding will be made as follows:
- For the first threshold we make a left thresholding. This eliminates all part of the histogram to the left of seuil_1.
- For the second seuil_2 we remove the part located on the right.

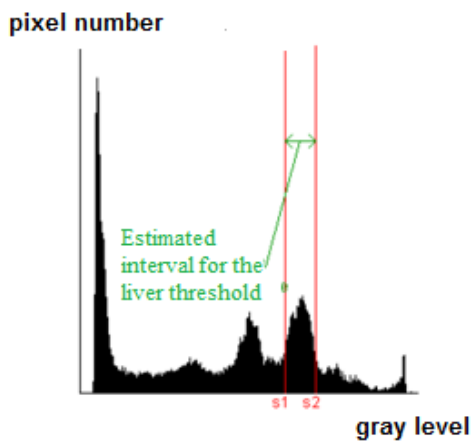

Figure 3: Detection of liver interval in the histogram

Canadian Journal on Image Processing & Computer Vision Vol. 2, No. 8, December 2011

This algorithm outputs a binary image in which several objects are presented in white including the liver, which is the largest among these objects. The resulting image has the disadvantage is that the pixels that represent a certain object are dispersed. This leads us to pose the problem to make them continuous so that the object is more clear and smoother. And to do this it is generally necessary to apply a filter.

The median filter is used to ameliorate the image. The morphological filter is applied to fill holes. The GCC algorithm is developed for the detection of the largest connected component. It contains two modules: the first for the labeling of connected components, the second for finding the largest connected component.

*num ← 0*
*for all P pixels from left to right and top to bottom:*
*if P is different from the background*
*then*
    *if only one of two top and left neighbors has a label e*
*then*
       *mark P with e*
      *else if these two neighbors have the same label*
        *then mark p with e*
         *else if then mark*
           *P with min(eti1, eti2)*
            *put (e1, e2) in table of*
           *equivalence*
          *else num ← num+1*
          *mark P with num*
         *end if*
        *end if*
*End if*

Algorithm 1: Labeling of connected components

*Let $e_1,...., e_n$ integers, $e_i$ label of the connected component number i.*
*-Nbr is the number of connected components.*
*-tab [nbr] is a table for storing the number of pixels of each component.*

*1. For each pixel p*
  *if it has a label $e_i$ then*
    *tab [$e_i$] ← tab [$e_i$] +1*
*end for*
*gcc ← index max (tab)*
*2. Browse image labels*
*if the current pixel p has a label gcc*
    *then display p in the resulting image*
*end if*
*end for*

Algorithm 2: Search for the largest component

Contour detection was performed by applying the nonlinear filter of Sobel. The superposition of the contour on the original image allows us to deduct the region of the liver.

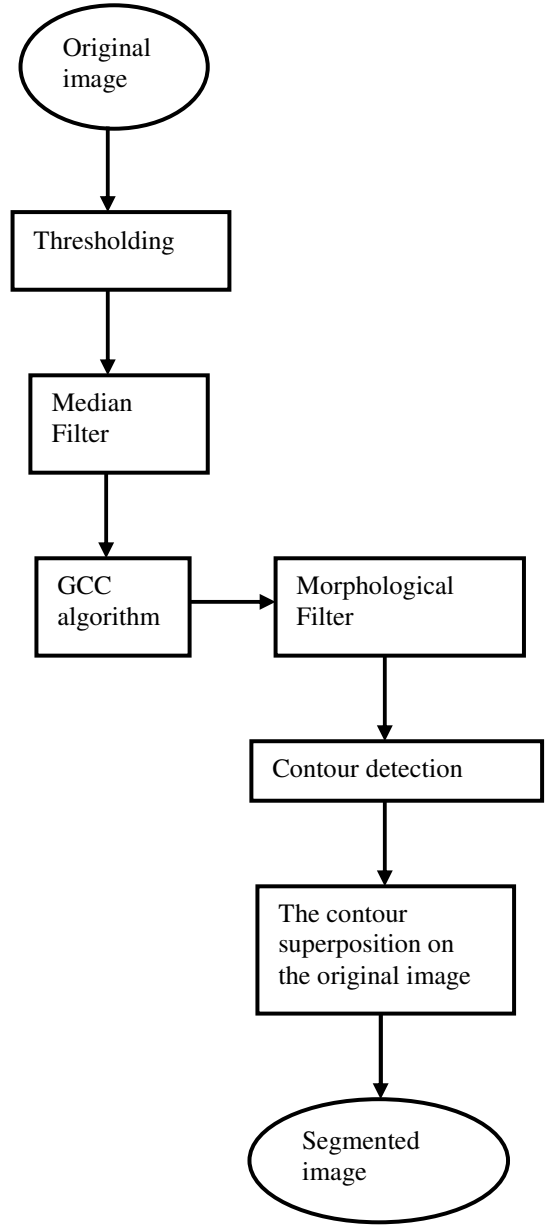

Figure 4: General structure of the proposed system

### 4. EXPERIMENTATION

Our segmentation method of liver was tested with ten abdomens (512x512 CT images) of ten patients. Each sample shows eight images that are the original image, the result of thresholding, application of median filter, the result of the detection of GCC, morphological closing, edge detection of the liver, the final result, manual segmentation of the expert. Figure 2 shows the seven images of a sample.

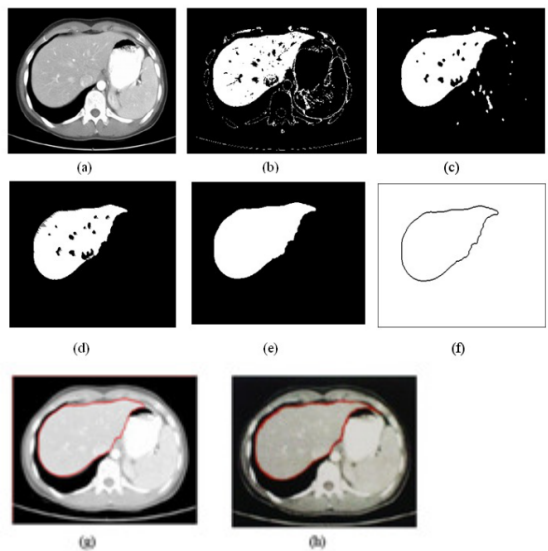

Figure 5: Good segmentation: example: (a) Original image, (b) binary thresholding, (c) Application of median filter, (d) Detection of the GCC, (e) morphological closing, (f) Edge detection (Sobel), (g) Final result, (h) manual segmentation of the expert.

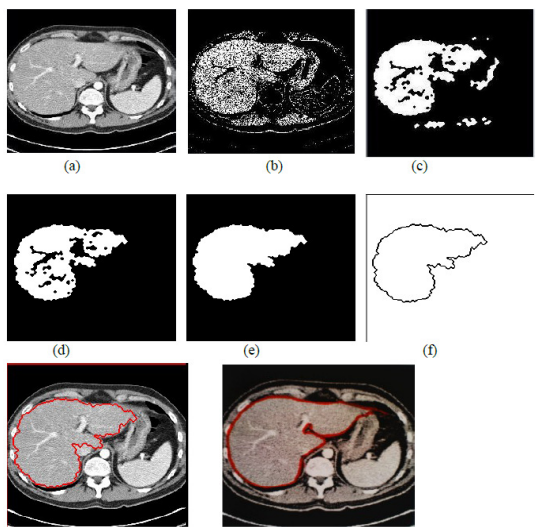

Figure 6: Average segmentation: example: (a) Original image, (b) binary thresholding, (c) Application of median filter, (d) Detection of the GCC, (e) morphological closing, (f) Edge detection (Sobel), (g) Final result, (h) manual segmentation of the expert.

In that case of execution, the result obtained by our system seems to be medium by report has the manual segmentation of the expert.

Apparently, it is impossible to obtain always a perfect result for the automatic segmentation. That's returns to various causes: in the images is different from a scanner to another, so the class of images is not the same and this poses a problem of automatic recognition as the nature of the histogram (the speed) depends strongly on the input image so the device with which the images are acquired. In some cases the peak representing the liver in the histogram is completely different compared to other images. And therefore the extraction of the liver is not satisfactory.

## 5. CONCLUSION

The results of this work are acceptable. In fact, we have tested our prototype on a CT scan images, allowing us to conclude that even our software is not one hundred percent reliable, and able to perform better. Indeed, we propose to improve our method by integrating the fuzzy logic to better manage inaccuracies.